\documentclass[conference,a4paper]{IEEEtran}
\IEEEoverridecommandlockouts

\usepackage[hidelinks]{hyperref}
\usepackage[cmex10]{amsmath}
\usepackage{amssymb,amsfonts}
\usepackage{dblfloatfix}

\usepackage[ruled,vlined]{algorithm2e}
\usepackage{graphicx}
\graphicspath{{Figures/PDF/}{Figures/PNG/}}

\usepackage{lineno}
\usepackage{booktabs}
\usepackage{siunitx}
\usepackage[numbers,compress]{natbib}
\usepackage{texnames}
\usepackage{bm,bbm}
\usepackage{orcidlink}
\usepackage{placeins}

\usepackage{booktabs} 
\usepackage{siunitx}  

\begin{document}

\title{\uppercase{ SAMST: A Transformer framework based on SAM pseudo label filtering for remote sensing semi-supervised semantic segmentation }
}


\author{
    \IEEEauthorblockN{
        Jun Yin$^{1}$, 
        Fei Wu$^{1}$,
        Yupeng Ren$^{2}$,
        Jisheng Huang$^{2}$,
        Qiankun Li$^{2}$,
        Heng jin$^{2}$,\\
        Jianhai Fu$^{2*}$\orcidlink{0009-0009-2819-3717}, 
        Chanjie Cui$^{2*}$\orcidlink{0009-0007-2220-0512}
    }
    \IEEEauthorblockA{
        Zhejiang University \\
        Zhejiang Dahua Technology Co., Ltd \\
        Correspondence: fujianhai2024@gmail.com (J.F.), cuichj@mail2.sysu.edu.cn (C.C.)
    }
}

\maketitle
\begin{abstract}
	Public remote sensing datasets often face limitations in universality due to resolution variability and inconsistent land cover category definitions. To harness the vast pool of unlabeled remote sensing data, we propose SAMST, a semi-supervised semantic segmentation method. SAMST leverages the strengths of the Segment Anything Model (SAM) in zero-shot generalization and boundary detection. SAMST iteratively refines pseudo-labels through two main components: supervised model self-training using both labeled and pseudo-labeled data, and a SAM-based Pseudo-label Refiner. The Pseudo-label Refiner comprises three modules: a Threshold Filter Module for preprocessing, a Prompt Generation Module for extracting connected regions and generating prompts for SAM, and a Label Refinement Module for final label stitching. By integrating the generalization power of large models with the training efficiency of small models, SAMST improves pseudo-label accuracy, thereby enhancing overall model performance. Experiments on the Potsdam dataset validate the effectiveness and feasibility of SAMST, demonstrating its potential to address the challenges posed by limited labeled data in remote sensing semantic segmentation.

\end{abstract}


\section{Introduction}

Semantic segmentation of remote sensing images is a crucial task in computer vision, facilitating applications such as land cover classification, urban planning, and environmental monitoring. Nevertheless, the scarcity of labeled data in this field presents a formidable challenge, given that pixel-level annotation is laborious and time-consuming. To tackle this issue, semi-supervised learning (SSL) methods have garnered significant attention, harnessing both labeled and unlabeled data to enhance model performance. Recent research primarily falls into two categories: consistency regularization (CR) ~\cite{10114409}~\cite{rs14040879}~\cite{9645575}~\cite{9566785}~\cite{8417973}~\cite{tarvainen2017mean} and self-training (ST) ~\cite{10542113}~\cite{10640951}~\cite{li2023semi}~\cite{cui2023semi}~\cite{yang2022st++}~\cite{lu2022simple}~\cite{li2021semisupervised}~\cite{wu2020semi}~\cite{ouali2020semi}~\cite{mittal2019semi}~\cite{lee2013pseudo}. CR ensures that consistent outputs are produced when different perturbations of the same data are provided as input. Self-training involves utilizing pseudo-labels produced by a model that has been trained on labeled data and the corresponding images to train the model further. Among these, pseudo-labeling techniques have garnered promise, with models generating labels for unlabeled data and iteratively refining them during training. However, existing pseudo-labeling methods often grapple with noisy labels.

The advent of foundation models, especially the Segment Anything Model (SAM)~\cite{kirillov2023segment}, has showcased impressive zero-shot generalization and boundary detection capabilities, rendering them highly suitable for overcoming these challenges. SAM's proficiency in segmenting objects with minimal prompts has proven effective in diverse domains, including medical imaging and remote sensing, where it has been incorporated into semi-supervised frameworks to boost pseudo-label accuracy  ~\cite{yang2024sam}. Nonetheless, directly applying SAM to remote sensing semantic segmentation is fraught with difficulties due to the distinctive attributes of remote sensing data, such as high-resolution variability and the demand for precise land cover delineation.

To address these challenges, we introduce SAMST, a novel semi-supervised semantic segmentation method designed specifically for remote sensing images. SAMST harnesses SAM's generalization capabilities while overcoming the limitations of existing pseudo-labeling techniques. Our approach comprises two core components: (1) a supervised model that undergoes self-training with both labeled and pseudo-labeled data, and (2) a SAM-based Pseudo-label Refiner. 

Our contributions can be summarized as follows:
\begin{itemize}
    \item We present SAMST, a semi-supervised semantic segmentation framework tailored for remote sensing images, tackling the issues of resolution variability and inconsistent land cover definitions.
    \item We develop a SAM-based Pseudo-label Refiner that iteratively refines pseudo-labels via threshold filtering, prompt generation, and label stitching, thereby improving the quality of pseudo-labels for unlabeled data.
    \item We introduce a composite weighted loss function to address the challenges of incorrect labels and class distribution shifts in pseudo-labeled data. This loss function combines labeled and pseudo-labeled losses, weighted differently to enhance model robustness.
    \item We showcase the effectiveness of SAMST on the Potsdam dataset, attaining state-of-the-art performance and demonstrating its applicability to remote sensing tasks.
\end{itemize}

\begin{figure}[t]
	\centering
	\includegraphics[width=0.5\textwidth]{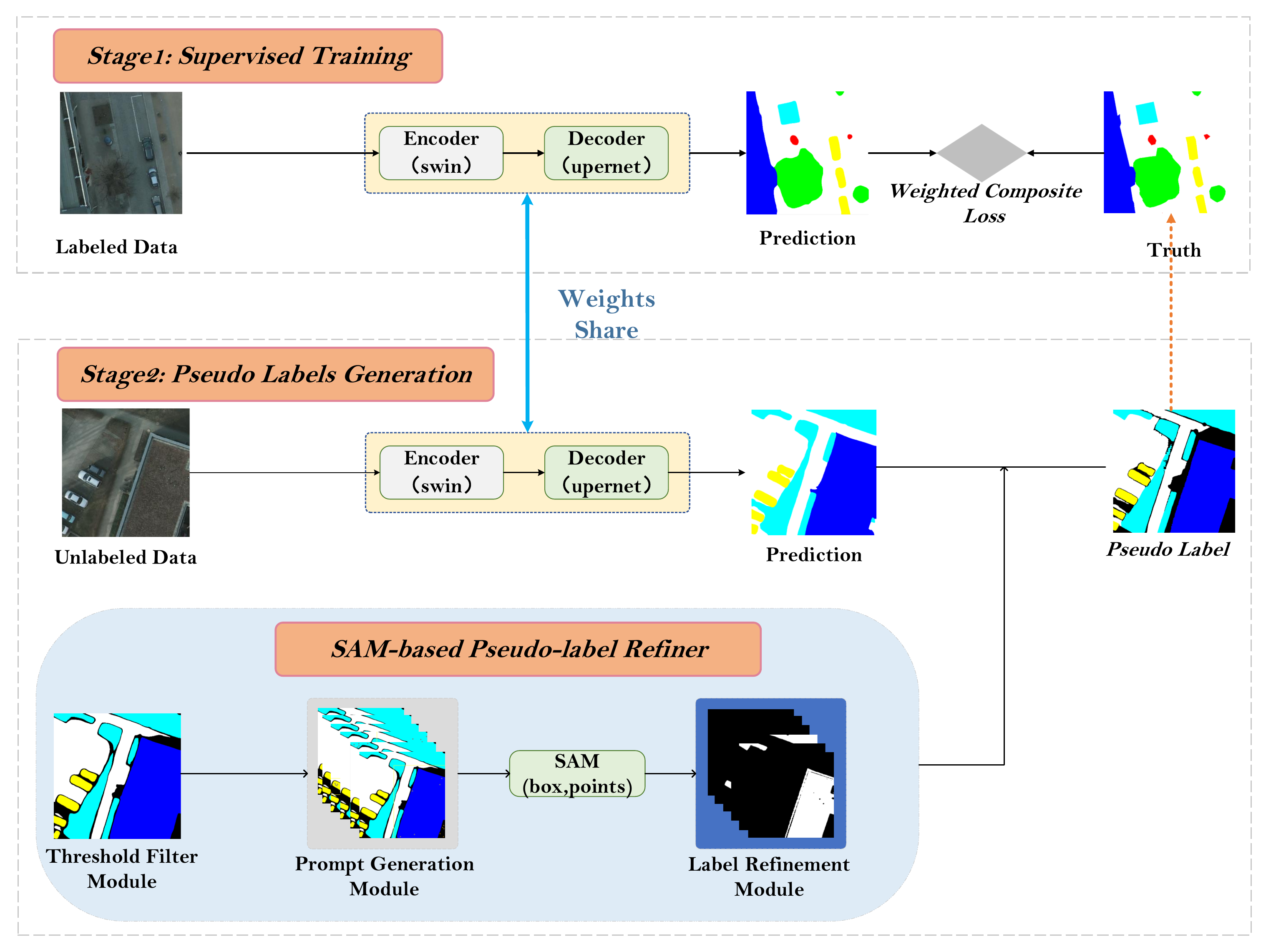}
	\caption{Overview of SAMST Framework with Two Stages: Supervised Training and Pseudo-label Generation}\label{fig:SAMST}
\end{figure}

\begin{table*}[!hbt]
    \centering
    \small
    \caption{Performance Comparision Study on the Postdam Dataset }\label{tab:comparison}
    \begin{tabular}{l S[table-format=2.2] S[table-format=2.2] S[table-format=2.2] S[table-format=2.2] S[table-format=2.2] S[table-format=2.2] S[table-format=2.2] S[table-format=2.2] S[table-format=2.2]}
        \toprule
        \multicolumn{4}{c}{\textbf{Overall Metrics (\%)}} & \multicolumn{6}{c}{\textbf{F1 per category (\%)}} \\
        \cmidrule(lr){1-4} \cmidrule(lr){5-10}
        \textbf{Method} & \textbf{mIoU} & \textbf{OA} & \textbf{mF1} & \textbf{Clutter} & \textbf{Car} & \textbf{Tree} & \textbf{Low Veg.} & \textbf{Building} & \textbf{Surface} \\
        \midrule
        Baseline & 67.00 & 93.48 & 78.56 & 46.43 & 87.76 & 76.03 & 80.74 & 92.61 & 87.83 \\
        LSST~\cite{lu2022simple} & 60.82 & 80.57 & 74.15 & 42.04 & 76.87 & 78.39 & 77.21 & 87.34 & 83.04 \\
        ST++~\cite{yang2022st++} & 64.89 & 83.83 & 75.37 & 23.32 & 88.18 & 82.93 & 81.64 & 89.39 & 86.78 \\
        ClassHyPer~\cite{rs14040879} & 66.58 & 83.62 & 78.23 & 43.55 & 85.71 & 82.69 & 80.42 & 90.62 & 86.40 \\
        SAMST (Ours) & 70.80 & 86.44 & 81.65 & 54.64 & 89.56 & 83.56 & 79.05 & 93.80 & 89.28 \\
        \bottomrule
    \end{tabular}
\end{table*}

\section{Method}
Fig.~\ref{fig:SAMST} illustrates the architecture of our proposed method, SAMST, which encompasses two distinct phases: supervised training and pseudo-label generation. In the initial supervised training phase, the model is trained on a combination of labeled and pseudo-labeled data. During the first iteration, only labeled data is used to initialize the model. Subsequently, in the pseudo-label generation phase, the pretrained model leverages the unlabeled data to generate pseudo-labels. These predictions undergo refinement using our SAM-based Pseudo-label Refiner module, thereby enhancing the quality of the pseudo-labels.

\subsection{SAM-based Pseudo-label Refiner}
Training an initial model on a limited labeled dataset often yields pseudo-labels with inadequate accuracy, potentially causing cumulative errors during iterative model training and misguiding the model's computation. To mitigate this, we design a SAM-based pseudo-label refinement module to enhance pseudo-label accuracy by minimizing incorrect label introduction and improving boundary precision. This module comprises three key components: the Threshold Filter Module (TFM), Prompt Generation Module (PGM), and Label Refinement Module (LRM).

\subsubsection{Threshold Filter Module}
Initially, the model trained on the first stage of supervised learning predicts unlabeled data. The predicted probability values are retained prior to applying argmax. These values are filtered using pre-set class-specific thresholds. Predictions meeting the threshold criteria are preserved, while others are marked as 255(indicating an ignored label).

\subsubsection{Prompt Generation Module}
\label{sec:pgm}

SAM operates as a prompt-based segmentation model, accepting prompt points and boxes for segmentation. we propose a Prompt Generation Module (PGM) to extract target boxes and positive/negative points for SAM. The PGM processes the output of the Threshold Filter Module (TFM) to generate precise segmentation contours by providing SAM with accurate prompts. The detailed process is described as follows:

First, for each class \( C_i \) (excluding \( C_i = 255 \)), all connected regions \( A_{C_i} \) are identified. For each connected region \( A_{C_i} \), the maximum bounding rectangle is computed and expanded by \( B_n \) pixels to form the prompt box \( B_{C_i} \) for SAM.

Second, positive points are randomly placed within the prompt box \( B_{C_i} \). The number of positive points is determined by \( P_p \). These points must satisfy the following conditions:
\begin{itemize}
    \item Located within the current connected region \( A_{C_i} \) with a predicted probability \( p_{i,j} \) exceeding the threshold \( T_p \):
    \begin{linenomath}
    \end{linenomath}
    \item All \( P_m \) surrounding pixels are within the connected region \( A_{C_i} \).
\end{itemize}

Third, negative points are randomly placed within the prompt box \( B_{C_i} \). The number of negative points is determined by \( P_n \). These points must satisfy the following conditions:
\begin{itemize}
    \item Located outside the current connected region \( A_{C_i} \) with a predicted probability \( p_{i,j} \) exceeding the threshold \( T_n \):
\begin{linenomath} 
    \end{linenomath}
    \item All \( N_m \) surrounding pixels are outside the connected region \( A_{C_i} \).
\end{itemize}

Finally, the parameters \( B_n \), \( P_p \), \( P_n \), \( T_p \), \( P_m \), \( T_n \), and \( N_m \) are determined based on the dataset and categories, with specific values set through experimental outcomes.

By following these steps, the PGM module provides precise target boxes and positive/negative points for SAM, enabling the generation of high-quality segmentation contours.

\subsubsection{Label Refinement Module}

The PGM feeds each set of prompt boxes and points into SAM to obtain masks for each connected region. The LRM concatenates these masks to form the updated pseudo-label. For each category's mask \( \text{mask}_m \), holes are first removed to obtain \( \text{mask}_{m1} \). It is then compared with the initial model's predicted probabilities. If the predicted category matches and the probability exceeds \( t_c \), or if it does not match but the non-matching category's probability is below \( t_o \), the pixel's pseudo-label is updated to the current category. Otherwise, it is set to 255. Here, \( t_c \) and \( t_o \) are thresholds for filtering.

\subsection{Weighted Composite Loss}
In the process of generating pseudo-labels, the introduction of incorrect labels is inevitable. Moreover, due to the model's varying recognition accuracy for different classes within the dataset, the class distribution of the pseudo-labeled dataset may differ from that of the original labeled dataset. To mitigate this issue, we propose a composite weighted loss function (1).

\begin{linenomath}
\begin{equation}
L_{\text{wc}} = L_{\text{label}} + \alpha L_{\text{pseudo}}
\end{equation}
\end{linenomath}

The composite weighted loss consists of two components: the labeled loss and the pseudo-labeled loss. The pseudo-labeled loss is weighted by a factor \(\alpha\). Both the labeled loss and pseudo-labeled loss are weighted cross-entropy loss functions, with different weights applied to each.

\begin{linenomath}
\begin{equation}
L_{\text{label}} = -\sum_{c=1}^{M} w_{\text{l}, c} y_c \log(p_c) \quad
\end{equation}
\end{linenomath}

\begin{linenomath}
\begin{equation}
L_{\text{pseudo}} = -\sum_{c=1}^{M} w_{\text{p}, c} y_c' \log(p_c) \quad
\end{equation}
\end{linenomath}

Where \(w_{\text{l}, c}\) and \(w_{\text{p}, c}\) are the class weights for the labeled and pseudo-labeled losses, respectively.

\section{Experiments}
\subsection{Datasets and Experimental Setup}
The Potsdam dataset~\cite{rottensteiner2012isprs} is frequently employed for remote sensing (RS) image segmentation tasks. It comprises high-resolution aerial images captured in Potsdam, Germany. The dataset contains 38 aerial images, each with a resolution of 5 cm/pixel and dimensions of 6000 × 6000 pixels. These images encompass a variety of scenes within and around Potsdam, such as urban areas, rivers, forests, grasslands, and farmland. Each pixel in the Potsdam dataset is annotated with one of six classes: Impervious Surface, Building, Low Vegetation, Tree, Car, and Clutter. Corresponding labels are provided for each pixel, facilitating the training and evaluation of RS image segmentation algorithms. The Potsdam dataset has been extensively utilized in research and competitions, including the ISPRS 2-D Semantic Labeling Contest. 

For our experiments, we allocated 24 images for training and 14 for validation and testing. To streamline the training process, we partitioned the images into non-overlapping patches measuring 512 × 512 pixels. Consequently, the training set comprises 3456 patches, whereas the testing set includes 2016 patches. Typically, only a small portion of the training set (1/32), equivalent to 108 patches, is utilized as labeled data. Unless otherwise stated, the remaining patches (3348) are considered unlabeled data.

We evaluate the performance of our methods using the F1 score, overall accuracy (OA), and Mean Intersection over Union (MIoU)~\cite{csurka2013good}.

\subsection{Implementation Details}
We implemented the experimental code using PyTorch. Our semi-supervised model employs the Swin transformer~\cite{liu2021swin} as the backbone and Uperhead ~\cite{xiao2018unified} as the primary segmentation head. Additionally, we use FCNhead~\cite{zhang2018context} as an auxiliary segmentation head.During training, we used the same hyperparameters for both the supervised training stage and the pseudo-label training stage. Training was conducted on eight A40 GPUs, with a batch size of 4 per GPU. We utilized the AdamW optimizer with an initial learning rate of 0.00006 and a weight decay of 0.01. The learning rate decay was linear, with a warm-up period of 1500 iterations.

We trained the semi-supervised model using 1/32 of the data as labeled data and all remaining data as unlabeled. Throughout the training process, we completed a total of 80,000 iterations, saving a checkpoint every 1,000 iterations. We used the best checkpoint to generate pseudo-labels, completing one full iteration.

\begin{figure}[!htb]
	\centering
	\includegraphics[width=0.5\textwidth]{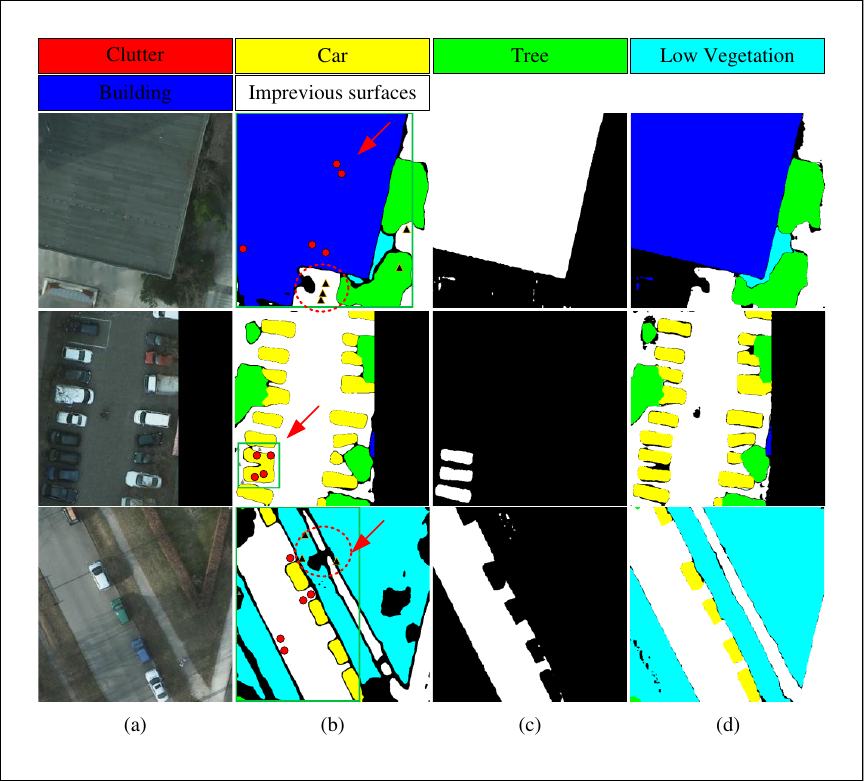}
	\caption{Visualization results of our proposed SAM-based Pseudo-label Refiner module. Panel (a) shows the input image. Panel (b) depicts the prompt boxes (green rectangles) and prompt points (red circles indicate positive points, black triangles indicate negative.  Panel (c) shows the segmentation results generated by SAM, while Panel (d) demonstrates the refined pseudo-label output.}\label{fig:SAM_result}
\end{figure}

\begin{figure}[!htb]
	\centering
	\includegraphics[width=0.5\textwidth]{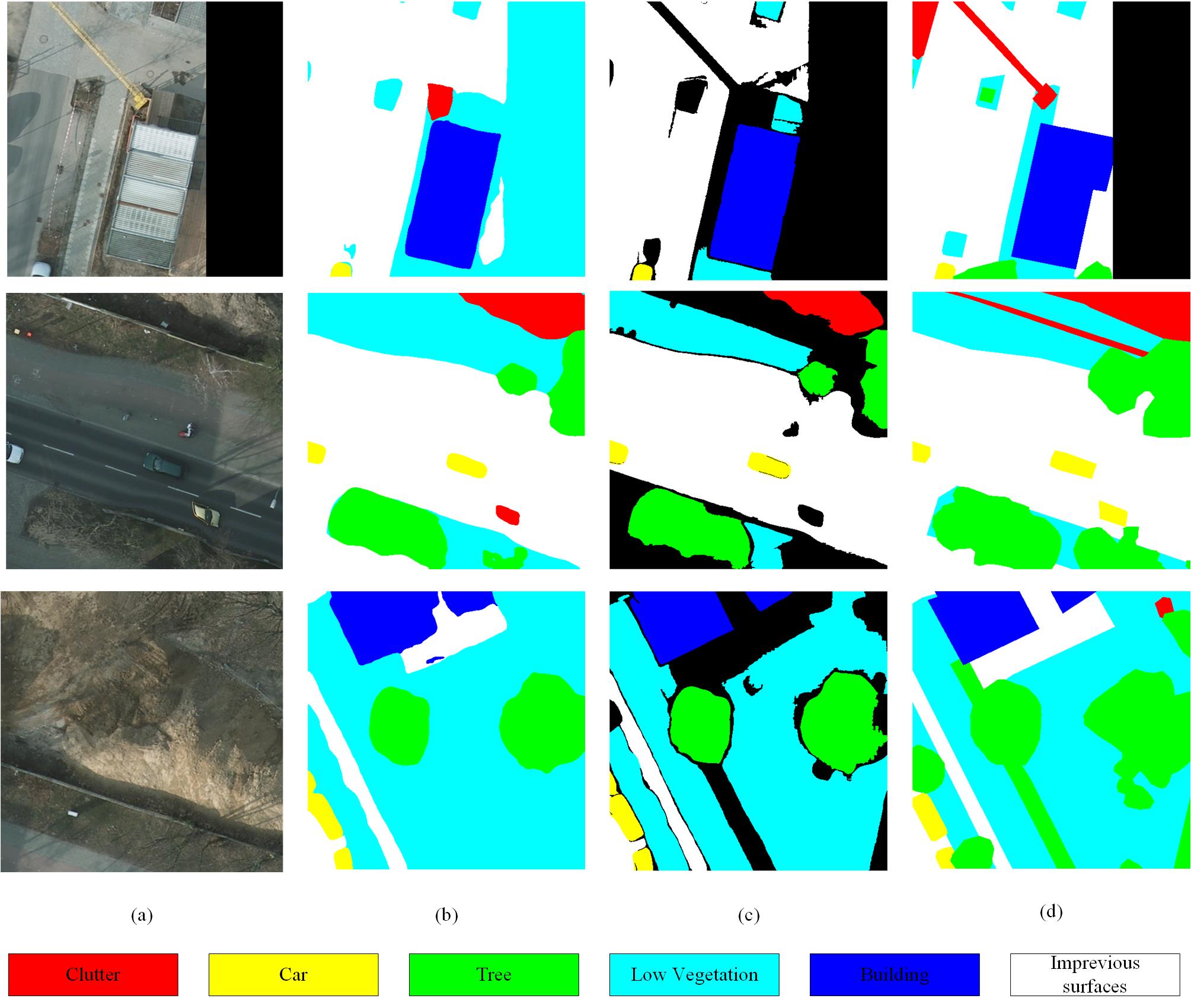}
	\caption{Visual Comparison of Pseudo Result with True Label. (a) depicts the input images. (b) shows the predictions from the initial supervised model. (c) presents the refined pseudo-labels generated by our SAM-based Pseudo-label Refiner module. (d) displays the true label.}
    \label{fig:pseudo_label}
\end{figure}

\subsection{Experimental Results}

Table~\ref{tab:comparison} presents a comprehensive comparative analysis of our method, which involves only a single iteration of SAMST, against several other semi-supervised approaches on the Potsdam test dataset, with a focus on a \(1/32\) labeled data ratio. The table clearly demonstrates significant improvements in various metrics for our method. Specifically, compared to the baseline method, we achieved enhancements of \(5.67\%\) in mIoU, \(2.11\%\) in overall accuracy (OA), and \(3.93\%\) in the average F1 score. On a per-class basis, after a single iteration of pseudo-label training, there was a notable increase in precision for most classes, with the exception of low vegetation, which saw a slight decrease. Notably, the F1 score for the clutter class improved by \(17.68\%\). Compared to methods based on LSST~\cite{lu2022simple}, ST++~\cite{yang2022st++}, and ClassHyper~\cite{rs14040879}, our method increased the mIoU by \(16.41\%\), \(9.11\%\), and \(6.34\%\), respectively.

Fig.~\ref{fig:SAM_result} shows the visualization results of our proposed SAM-based Pseudo-label Refiner module. This figure highlights three key capabilities of our module: (1) Boundary refinement: As shown in the second row, the originally adherent car masks are effectively separated into distinct instances. (2) Error filtering: Incorrect masks are eliminated. For example, building misclassified within the prompt box (bottom-left, first row) are correctly reassigned to the invalid class (label 255) after refinement. (3) Boundary completion: Incomplete object boundaries are reconstructed, as evidenced by the restoration of impervious surfaces (top-left, third row).

\begin{figure}[!htb]
	\centering
	\includegraphics[width=0.5\textwidth]{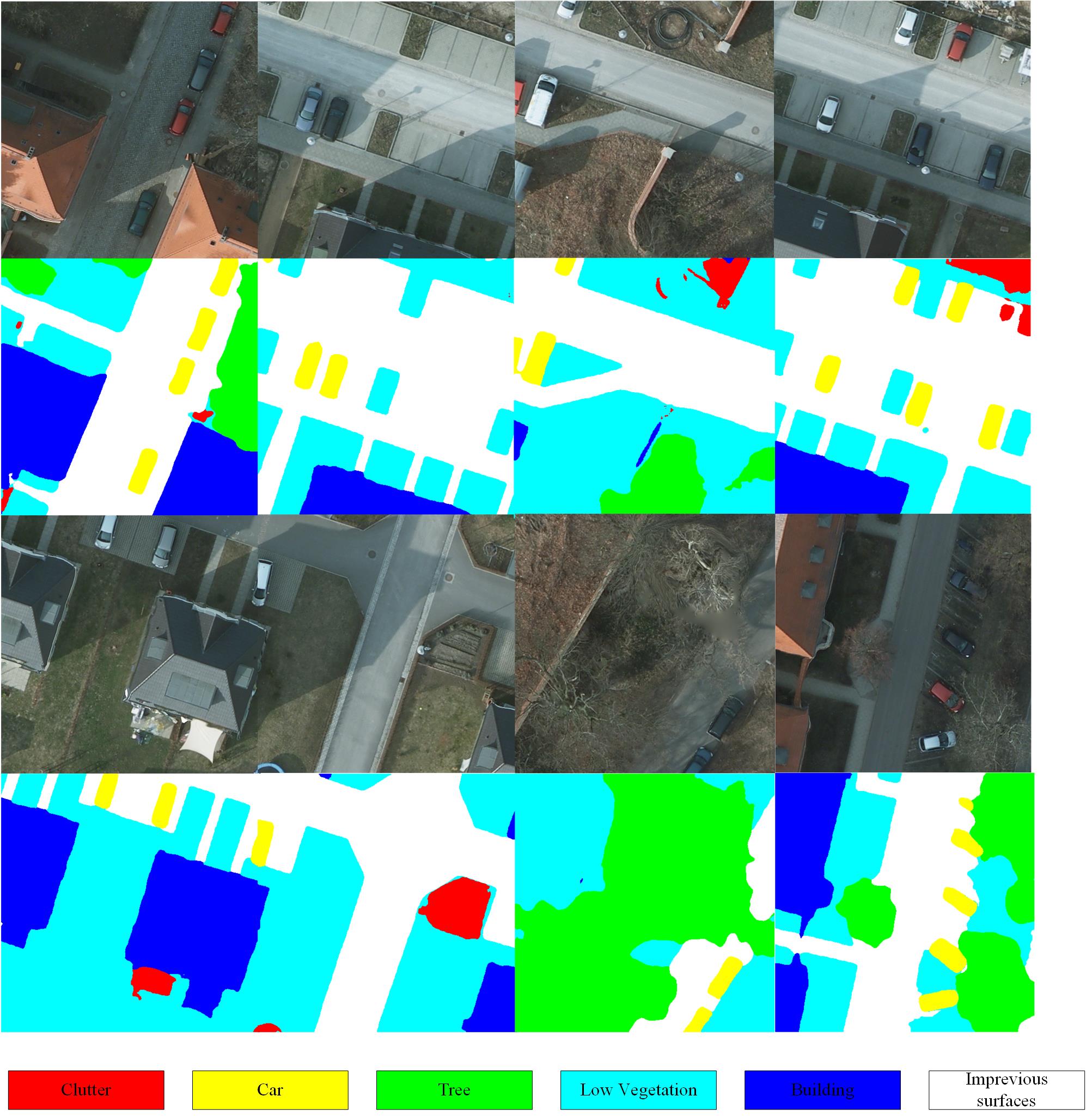}
	\caption{Visual Result with our proposed SAMST framework after one iteration.}\label{fig:result}
\end{figure}

Fig.~\ref{fig:pseudo_label} illustrates the comparison of pseudo-label results generated by our SAMST framework using labeled data excluded from the training phase. In the first row, the initial model's prediction (b) inaccurately omits and contours a tower crane, classified as clutter. Our SAM-based Pseudo-label Refiner module corrects this, accurately segmenting and labeling the tower crane as 255. In the second row, cars initially misclassified as clutter, with clutter areas obscured by low vegetation, are correctly identified and labeled as 255 by our module. In the third row, building contours are optimized, and misclassified low vegetation adjacent to trees is corrected to 255. Overall, our SAM-based Pseudo-label Refiner module effectively minimizes prediction errors in pseudo-labels. Fig.~\ref{fig:result} shows the results of our proposed SAMST method after one iteration.

\section{Conclusion}
In summary, this paper introduces SAMST, a novel semi-supervised semantic segmentation method for remote sensing images. A key innovation is the SAM-based Pseudo-label Refiner, which utilizes SAM's robust generalization to eliminate erroneous labels and refine boundaries in pseudo-labels. Furthermore, the Label Refinement Module addresses semantic discrepancies between SAM and remote sensing datasets, ensuring accurate category alignment and preventing the introduction of new errors. These contributions enhance the precision of pseudo-labels and overall segmentation performance, as evidenced by state-of-the-art results on the Potsdam dataset. Despite these advancements, the current approach includes dataset-specific hyperparameters that require further optimization. Future work will aim to refine these parameters and expand the applicability of SAMST.

\small
\bibliographystyle{IEEEtranN}
\bibliography{IGARSS2025LaTeXTemplate/reference_SAMST}

\end{document}